\documentclass[manuscript,screen]{acmart}
\usepackage{graphicx}
\usepackage{color,soul}
\usepackage{graphicx}
\usepackage[english]{babel}
\usepackage{comment}
\usepackage{hyperref}

\usepackage{color, colortbl}
	
\definecolor{blue}{rgb}{0.8,0.9,1}

\AtBeginDocument{%
  \providecommand\BibTeX{{%
    \normalfont B\kern-0.5em{\scshape i\kern-0.25em b}\kern-0.8em\TeX}}}

\setcopyright{acmcopyright}
\copyrightyear{2022}
\acmYear{2022}

\acmConference[ACM Computing Surveys Journal]{ACM Computing Surveys Journal}{December 12--24, 2022}

\begin{document}

\title{The Recent Advances in Automatic Term Extraction: A survey}

\author{Hanh Thi Hong Tran}
\email{hanh.tran@ijs.si}
\orcid{0000-0002-5993-1630}
\affiliation{%
  \institution{Jozef Stefan Institute, Slovenia; University of La Rochelle, France}
  \streetaddress{Jamova cesta 39}
  \city{Ljubljana}
  \country{Slovenia}
  \postcode{1000}
}

\author{Matej Martinc}
\email{matej.martinc@ijs.si}
\orcid{0000-0002-7384-8112}
\affiliation{%
  \institution{Jozef Stefan Institute, Slovenia}
  \streetaddress{Jamova cesta 39}
  \city{Ljubljana}
  \country{Slovenia}
  \postcode{1000}
}

\author{Jaya Caporusso}
\email{jaya.caporusso96@gmail.com}
\affiliation{%
  \institution{Jozef Stefan Institute, Slovenia}
  \streetaddress{Jamova cesta 39}
  \city{Ljubljana}
  \country{Slovenia}
  \postcode{1000}
}

\author{Antoine Doucet}
\orcid{0000-0001-6160-3356}
\affiliation{%
  \institution{University of La Rochelle, France}
  \streetaddress{23 Av. Albert Einstein}
  \city{La Rochelle}
  \country{France}
  \postcode{17000}
}

\author{Senja Pollak}
\email{senja.pollak@ijs.si}
\orcid{0000-0002-4380-0863}
\affiliation{%
  \institution{Jozef Stefan Institute, Slovenia}
  \streetaddress{Jamova cesta 39}
  \city{Ljubljana}
  \country{Slovenia}
  \postcode{1000}
}

\renewcommand{\shortauthors}{Tran et al.}


\begin{abstract}
Automatic term extraction (ATE) is a Natural Language Processing (NLP) task that eases the effort of manually identifying terms from domain-specific corpora by providing a list of candidate terms. As units of knowledge in a specific field of expertise, extracted terms are not only beneficial for several terminographical tasks, but also support and improve several complex downstream tasks, e.g., information retrieval, machine translation, topic detection, and sentiment analysis. ATE systems, along with annotated datasets, have been studied and developed widely for decades, but recently we observed a surge in novel neural systems for the task at hand. Despite a large amount of new research on ATE, systematic survey studies covering novel neural approaches are lacking. We present a comprehensive survey of deep learning-based approaches to ATE, with a focus on Transformer-based neural models. The study also offers a comparison between these systems and previous ATE approaches, which were based on feature engineering and non-neural supervised learning algorithms. 
\end{abstract}





\maketitle


\section{Introduction}
Terms are textual expressions that denote concepts in a specific field of expertise. They are beneficial for several terminographical tasks performed by linguists, e.g., construction of term dictionaries \cite{le2010automating}. Moreover, terms can also improve several complex downstream natural language processing (NLP) tasks, e.g., information retrieval \cite{lingpeng2005improving} and machine translation \cite{wolf2011statistical}. Manual term extraction is labor- and time-demanding. Therefore, with the new advances in machine learning and the recent introduction of novel neural architectures, we observed a push towards automatizing the task and reducing the time and effort needed to identify terms in domain-specific corpora.

ATE systems usually employ a two-step procedure: (1) extracting a list of candidate terms, and (2) determining which candidate terms are correct using supervised or unsupervised techniques. Early ATE systems were based on handcrafted rules and substantial feature engineering effort, which relied on linguistic knowledge and distinctive linguistic aspects of terms. They employed several NLP tools, e.g., tokenization, lemmatization, stemming, chunking, and POS tagging. Next, several other studies opted for a statistical approach 
based on various measures, such as termhood \cite{vintar2010bilingual}, unithood \cite{daille1994towards}, or C-value \cite{frantzi1998c}. These strategies are also employed in some current hybrid systems, which rely on combining linguistic and statistical information.

Inspired by the success of deep neural networks on several NLP tasks, a variety of neural approaches were proposed for ATE over the past few years, the most popular being the approaches that rely on Transformer-based language models \cite{vaswani2017attention}. Such models are appealing due to two major reasons: (1) they can deduce features themselves, which dismisses the need for hand-crafted feature engineering; and (2) they typically do not require domain-specific resources and are thus more domain-independent. Overall, several studies used neural architectures for ATE, by either leveraging embedding representations as the input for the classifiers  \cite{amjadian2016local,zhang2018semre,kucza2018term,gao2019feature} or by employing pretrained language models followed by fine-tuning in the recent NLP transfer learning paradigm  \cite{rigouts2020termeval,conneau2020unsupervised,hazem2020termeval,trantransformer,tran2022can}. 

The recent advances in supervised machine learning and deep learning approaches have become more popular than the unsupervised ones in the context of term extraction. However, while the traditional approaches had been comprehensively addressed in previous surveys \cite{kageura1996methods,da2014survey}, no systematic summary of the improvements made in supervised term extraction systems using machine learning or deep learning has been proposed yet. Therefore, our paper aims to provide a comprehensive research trajectory in the supervised term extraction task, including the following aspects.
\begin{enumerate}
    \item We survey the resources for term extraction of the past three decades. All the well-annotated corpora are taken into consideration if the two following requirements are satisfied: (1) they are publicly available, and (2) a description paper explaining their construction for the term extraction was published;
    \item  We propose a systematic review of deep learning-based approaches to the term extraction task (with a specific focus on the Transformer-based models) and compare them with previous approaches based on feature engineering and shallow supervised learning algorithms. Our comparison analysis highlights the improvements achieved by neural networks and shows that incorporating some of the insights learned from past work on feature engineering-based ATE systems can yield further improvements; 
    \item  We present all the metrics used in term extraction and categorize them based on either indirect (i.e., through a downstream application) or direct evaluation methodology (i.e., whether the evaluation is human judges, dictionary-based, or gold standard based) and the scope of results (i.e., whether we evaluate the entire results, parts of results, or top-k best results);
    \item  Incorporating the lessons learned from the existing techniques, we point out the current challenges in correctly capturing multi-word and nested terms, and in improving the robustness of the term extractor. Based on the current challenges, we propose some suggestions for further exploration in the term extraction research.
\end{enumerate}

The rest of the survey is organized as follows. Section \ref{sec:survey} presents the related surveys of approaches to ATE written in the last few decades. In Section \ref{sec:method}, we introduce our approach to identifying the articles for the survey. Section \ref{sec:dataset} presents a comparative study of different monolingual and multilingual corpora that can be used for the training and evaluation of ATE systems. This is followed by a systematic description of different ATE approaches in Section \ref{sec:systems}, different evaluation metrics used in ATE in Section \ref{sec:eval}, and a comparative evaluation of different systems in Section \ref{sec:results}. Based on the current works and the trending technology, we discuss the challenges we are still facing and the possible future directions in Section \ref{sec:discussion}. Finally, we conclude our work in Section \ref{sec:conclusion}.

\section{Previous surveys and comparative studies}
\label{sec:survey}
The first comprehensive survey of approaches to solving ATE was written by \cite{kageura1996methods}. It provided a systematic overview of the two major trends in the principles and methods of automatic term recognition and introduced a distinction between linguistic and statistical approaches. \cite{pazienza2005terminology} argued that all contemporary algorithms employed linguistic methods as a filtering step. A decade later, \cite{da2014survey} presented a survey of different approaches in automatic term extraction, which can be divided into three sub-categories: statistical, linguistics, and hybrid (i.e., statistical and linguistic). Furthermore, they made a brief review of different corpora in different domains that have been used in term extraction, but only for Brazil Portuguese. Regarding monolingual rich-resourced languages such as English, \cite{astrakhantsev2015methods} presented a survey of existing notions of a term and its linguistic features. They formulated the definition of automatic term extraction, analyzed available approaches, and proposed a general ATE pipeline consisting of four consecutive steps: preprocessing, term candidate collecting, term candidate scoring, and term candidate ranking. 

Besides these initial survey studies, the systematic comparison of various ATE systems on the same corpora was also conducted in the scope of several shared tasks and workshops. NTCIR \cite{kageura2000ir} is an evaluation-based project for information retrieval and term extraction carried out between 1998 and 1999. The term recognition part of this project tackled three subtasks, including the initial term extraction, keyword extraction, and role analysis tasks. However, this study suffered from a limited number of participants and the absence of previous evaluation initiatives for computational terminology. Later, CoRReCT \cite{enguehard2003correct} introduced an interesting data set with two proposed systems to detect the candidate terms (i.e., FASTR  and SYRETE) and a new protocol to evaluate term detection, the so-called controlled indexing. A few years later, the Campagne d’Evaluation des Systèmes d’Acquisition des Ressources Terminologiques (CESART) \cite{el2006terminological}, presented term extraction as one of three subtasks with a more user-oriented evaluation. This project proposed an interesting new protocol for term extraction, including a gold standard list of terms and a corresponding domain-specific acquisition corpus. This setting is considered as one of the better baseline approaches toward term annotation in the latest term extraction studies.

The TermEval 2020 shared task on monolingual automatic term extraction, organized as part of the CompuTerm workshop \cite{rigouts2020termeval}, presented one of the first opportunities to systematically study and compare various term extraction systems on comparable corpora in English, French, and Dutch from four domains (Corruption, Wind, Equitation, Heart Failure). The participating systems ranged from traditional approaches based on linguistic and statistical features to systems based on shallow machine learning and neural networks. While the workshop was an important step forward in terms of systematic comparison among different methodologies, there is still room for improvement as the open-source code is not available for most participating systems, hindering their replicability.

As of yet, to the best of our knowledge, there is no comprehensive survey of modern neural network ATE systems, nor is there a survey that compares systems based on extensive feature engineering and neural systems in both multi-lingual and multi-domain settings. Recently proposed approaches, which in most cases rely on transferring knowledge from one domain to another or from one language to another, have not been thoroughly evaluated and a comparison between these systems and other ATE systems is still lacking.

\section{Methodology}
\label{sec:method}

\begin{figure}[ht]
    \centering
    \includegraphics[scale=0.38]{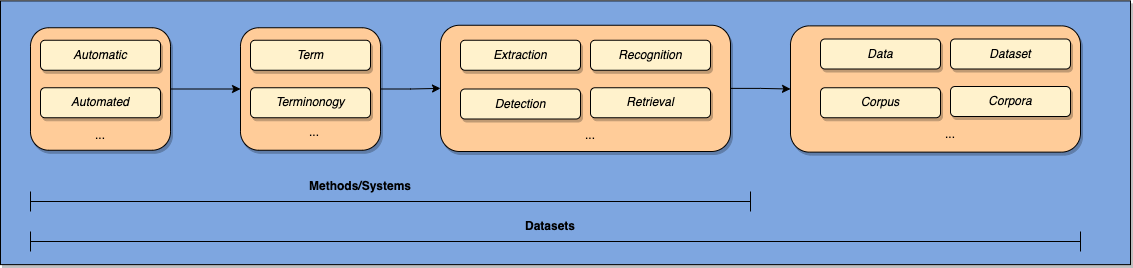}
    \caption{The combination of keywords to search for term datasets and systems.}
    \label{fig:search}
\end{figure}

We searched mainly on the five websites, including Google\footnote{https://www.google.com/}, Google Scholar\footnote{https://scholar.google.com/}, Semantic Scholar\footnote{https://www.semanticscholar.org/}, ACL Anthology\footnote{https://aclanthology.org/}, and Terminology\footnote{https://benjamins.com/catalog/term} journals to identify suitable articles for this survey. The search progress was taken from October to December 2022. Our query terms covered two distinct ATE resources: available corpora and neural systems. Therefore, we used 80 different queries to collect the papers that describe either methods/systems (16) or corpora (64), which are the combinations of different keywords as illustrated in Fig. \ref{fig:search} and are listed as examples below:
\begin{itemize}
    \item Methods/Systems: \textit{automatic term extraction, automatic term recognition, automatic term detection, automatic term retrieval, automated term extraction, automated term recognition, automated term detection, automated term retrieval,  automatic terminology extraction, automatic terminology recognition, automatic terminology detection, automatic terminology retrieval, automated terminology extraction, automated terminology recognition, automated terminology detection, automated terminology retrieval,...}
    \item Corpora: We combined the above keywords with each of the four following keywords: \textit{data, dataset, corpus, corpora}. For example, \textit{automatic term extraction data, automatic term extraction dataset, automatic term extraction corpus, automatic term extraction corpora}, to mention a few.
\end{itemize}

We sorted the resulting papers obtained by each query according to the citation counts. After collecting based on queries, we filtered out all the papers that have less than five citations if the papers were released before 2022. After this filtering step, we continued to select only the papers that either introduced a neural architecture for automatic term extraction or described a novel ATE dataset to investigate in our survey. There were no limits in the languages and the domains of the dataset that the papers aimed to solve the tasks.
When it came to articles presenting neural architecture, we included them in the survey only if they represented the first article to introduce it; otherwise, we traced citations back until we found the original paper that introduced the architecture. We followed the same citation-tracing approach when searching for ATE systems that employ extensive feature engineering for performance comparison. In total, 329 articles were collected, 139 articles were reviewed after filtering, and 112 articles were finally selected from the point of view of the dataset, code, and tool availability after applying the citation-tracing mechanism for the survey.

\section{ATE datasets}
\label{sec:dataset}

In this section, we cover linguistic resources for term extraction developed in the past three decades. More specifically, we present and analyze monolingual and multilingual domain-specific gold standard corpora that can be used for the training and evaluation phase of ATE systems. We only consider resources that are publicly available and for which a description paper explaining their construction has been published.

\subsection{Monolingual corpora}

Table \ref{tab:mono_data} lists manually annotated monolingual corpora that: (1) have been used for monolingual ATE evaluation, (2) contain more than 100 terms, (3) have annotation or gold standards, and (4) are publicly available at a specific URL or easily to reconstruct based on the data description. Note that the years on the Table refer to when the corpora were released and that the statistics are up-to-date with the latest release. We attach the link of each corpus in the footnote for facilitating access.

GENIA \cite{kim2003genia} is one of the best-known biomedical English corpora that was used for evaluating monolingual term extraction in many studies \cite{krauthammer2004term,nenadic2006mining,fedorenko2013broad,yuan2017supervised,kucza2018term,zhang2018adapted}. The version 3.0 of GENIA was a collection of 2000 abstracts in the articles extracted from MEDLINE database.  93,293 out of all 436,967 words were annotated as biological terms. This corpus was the first publicly available corpus with linguistically rich annotations that include sentence boundaries, term boundaries, term classifications, semi-structured coordinated clauses, recovered ellipsis in terms, etc.

Another example of a bio-text-mining dataset is the Colorado Richly Annotated Full Text Corpus (CRAFT) \cite{bada2012concept}. CRAFT is a collection of full-text, open-access biomedical journal articles, each of which is a member of the PubMed Central Open Access Subset\footnote{\url{https://www.ncbi.nlm.nih.gov/pmc/tools/openftlist/}}.  The latest public release (version 2.0) includes around 100,000 ontologies/terms annotated from 67 out of the 97 articles in seven different categories: Chemical Entities of Biological Interest, Cell Ontology, Entrez Gene, three sub-ontologies of Gene Ontology, NCBI Taxonomy, Protein Ontology, and Sequence Ontology. Further details are described in CRAFT website\footnote{\url{https://bionlp-corpora.sourceforge.net/CRAFT/}}. However, both GENIA and CRAFT consisted of applying ontologies on text, and not of labeling terms as an open task.

In the context of the Gene Ontology project\footnote{\url{http://www.geneontology.org/}}, the Gene Ontology corpus \cite{gene2004gene} provides structured controlled vocabularies of molecular and cellular biology that are freely available for community use in the annotation of genes, gene products, and sequences. Between 2005 and 2022,  a total of 7,694,564 annotations was made, with 43,329 valid terms, 4.024 obsoleted terms, and 2,438 merged terms. These terms were organized in a hierarchical manner under three sub-ontologies, including Molecular Function, Biological Process, and Cellular Component. Further statistical details about different versions from the beginning of 2018 up to October 2022 can be accessed via the Gene Ontology website\footnote{\url{http://geneontology.org/stats.html}}. A few approaches to term extraction have been tested on this corpus, for example, \cite{zheng2006identifying,daraselia2007automatic}.
 
The ACL Reference Dataset for Terminology Extraction and Classification (ACL RD-TEC) is a dataset for evaluating the term extraction and classification from literature in the domain of computational linguistics. Derived from the ACL Anthology Reference Corpus, ACL RD-TEC consists of two versions: ACL RD-TEC v1.0 and ACL RD-TEC v2.0. The 1.0 version \cite{qasemizadeh2014acl} was released in 2014, providing 82,000 manually annotated candidate terms of three types: valid, invalid, or technology terms. Successive to the ACL RD-TEC v1.0, the ACL RD-TEC v2.0 \cite{qasemizadeh2016acl} embraces 300 unique abstracts from articles in the ACL Anthology Reference Corpus, published between 1978 and 2006. The annotated terms are classified into one of the seven categories: Method, Tool, Language resource, Language resource product, Model, Measures and measurements, and Other. Both versions are experimented with and evaluated by several approaches, such as linguistic and statistical ones \cite{wang2016featureless}, machine learning ones \cite{wang2016featureless,maldonado2016self,hatty2017evaluating,yuan2017supervised}, or even deep learning ones \cite{kucza2018term,gromann2021multilingual,lang2021transforming}.

Regarding the Brazilian Portuguese language, the number of available Brazilian Portuguese corpora has increased significantly in different domains, including and excluding well-constructed gold standards. In our survey, we focus only on the accessible corpora that have well-annotated labels or gold standards. For example,  the Ceramic Coating corpus \cite{teline2004avaliaccao} collected texts (448,352 words in total) from 164 papers in the Industrial Ceramics Magazine. Each paper contains 4 to 8 pages on average (approximately 4,000 words per paper). In the domain of pediatrics,  JPED \cite{coulthard2005application} includes 283 texts from the Pediatrics Journal (Jornal de Pediatria) and a gold standard of 1,534 bigrams and 2,647 trigrams. Another popular corpus is ECO \cite{zavaglia2005avaliaccao}. This available corpus, constructed in the context of the BLOC-Eco project, contains 390 Portuguese documents pertaining to the domain of Ecology. In the context of Nanoscience and Nanotechnology, the N\&N \cite{di2008compilaccao,coleti2009primeiro} corpus includes 1,057 texts, divisible into 5 categories: scientific, divulgation-scientific, informative, technical-administrative, and others. EaD \cite{souza2010exercicio} contains 347 texts extracted from the Internet in the field of Distance Education. Further information about all the Brazilian Portuguese-related corpora is described in \cite{da2014survey}'s work.


\begin{table}[ht]
    \caption{List of monolingual term datasets.}
    \label{tab:mono_data}
    \centering

    \begin{tabular}{ccccccc}
    \toprule
    \textbf{Datasets} &  \textbf{Years}  & \textbf{Domain(s)} & \textbf{Language(s)}  & \textbf{No of docs} & \textbf{No of words} & \textbf{No of terms} \\  \midrule
    GENIA\footnote{\url{http://www.geniaproject.org/genia-corpus}} & 2003 & Biomedicine & English &  193 & 436,967 & 93,293 \\ \midrule
    GO\footnote{\url{http://geneontology.org/}}   & 2005 &  Gene  &English & - & - & 43,329 \\ \midrule
    CRAFT v2.0\footnote{\url{https://bionlp-corpora.sourceforge.net/CRAFT/}} & 2016 &  Biomedicine  & English & 67 & $>$ 560,000 & $\approx$ 100,000 \\ \midrule
     ACL  RD-TEC  v1.0 \footnote{\url{http://pars.ie/lr/acl-rd-tec-terminology}} & 2014 & Computational   & English &10,922 & 36,729,513 & 82,000 \\
     &  & linguistics &  & &  &  \\ \midrule
    ACL  RD-TEC  v2.0\footnote{\url{http://pars.ie/lr/acl_rd-tec}} &2016 &Computational  &English&300 &33,216 & 6,818 \\
     &  & linguistics &  & &  &  \\ \midrule
    JPED \footnote{\url{https://repositorio.ufsc.br/handle/123456789/102257}} & 2005 & Pediatrics  &Portuguese &  283  &-&- \\ \midrule 
    ECO \footnote{\url{http://www.nilc.icmc.usp.br/nilc/projects/bloc-eco.htm}} & 2005 &  Ecology &Portuguese & 390 &-& 322  \\ \midrule
    N\&N \footnote{\url{}}& 2008 & Nanoscience &  Portuguese & 1,057 & - & 1,794 \\
    &  & Nanotechnology   &   & &  & \\ \midrule 
    EaD \footnote{\url{}}& 2010 &  Distance  &Portuguese & 347 &-& 118\\
    &  &  education  & &  && \\ \midrule  
    Irish Wiki\footnote{\url{https://github.com/jmccrae/irish_saffron}} & 2019 & Education & Irish & 11 & 5,178 & 864 \\ \midrule
    Hindu Wiki\footnote{https://hi.wikipedia.org/w/api.php} & 2022 & Education & Hindu & 71 & 11,960 & 953 \\ \midrule
    RSDO5 \footnote{\url{https://www.clarin.si/repository/xmlui/handle/11356/1400}} & 2021& Ph.D. theses,   & Slovene & 12 & 257,029 & 37,985  \\
     & &   Scientific book,  &  &  &  &  \\
     & &   Journal articles,  &  &  &  &  \\
     & &   Graduate textbooks  &  &  &  &  \\
    
    \bottomrule
    \end{tabular}
    \label{tab:mono-dataset}
\end{table}

\newpage
For what concerns less-resourced languages, RSDO5\footnote{\url{https://www.clarin.si/repository/xmlui/handle/11356/1400}} is one of the academic Slovenian corpora with well-documented annotation guidelines. It consists of 12 texts, collected between 2000 and 2019, from the fields of biomechanics, linguistics, chemistry, and veterinary science. Over 250,000 words and almost 38,000 manually annotated terms were included. Apart from the manually annotated terms, the corpus was also annotated using the Universal Dependencies annotations (e.g., tokenization, sentence segmentation, lemmatization, morphological features, and dependency syntax). The corpus can serve as training sets for supervised term extraction and for term extraction tool benchmarking. Recently, while \cite{mccrae2019adapting} applied term extraction on 11  Wikipedia Irish documents of 5,178 words and 864 extracted terms, \cite{banerjee2022dataset} introduced a collection of 71 Wikipedia Hindu documents comprising 11,960 words and 953 manually annotated terms. 

The monolingual datasets differ considerably, as there is a lot of variation in both the annotation and evaluation. Diverse approaches have been applied to each different aspect, including annotation type (candidate term list or just source text, etc.), annotation scheme (binary or multi-label, etc.), annotation guidelines (term length, POS patterns, whether or not to add named entities, etc.). The inter-annotator agreement (IAA), a measure of how well multiple annotators can make the same annotation decision for a certain category, varies as well, and there is often a lack of meta-information to better understand the corpus. Another problem is that each monolingual dataset covers a different domain and usually covers only a limited amount of terms within that specific domain. Consequently, it is extremely difficult to make any comprehensive comparison among corpora or combine multiple datasets into an extensive corpus.

\subsection{Multilingual corpora}
With the development of multilingual tasks \cite{devlin2018bert,lang2021transforming}, datasets that cover multiple languages and domains are becoming increasingly important for the training and evaluation of multilingual and cross-lingual models. Table \ref{tab:multi-data} presents the available manually annotated corpora used in multilingual ATE training  and/or evaluation that cover more than one language.


The Terminology Extraction, Translation Tools and Comparable Corpora (TTC) project \cite{daille2012building,gornostay2012terminology} was one of the first domain-specific corpora made purposefully to enable training and evaluation of multilingual term extraction models. The corpus contains manually annotated texts from two domains (Wind energy and Mobile technology) in seven languages (Chinese, English, French, German, Latvian, Russian, and Spanish). The summary for each domain can be found on this website\footnote{\url{http://www.ttc-project.eu/}}.

\begin{table}[ht]
    \caption{A list of multilingual term datasets.}
    \label{tab:multi-data}
    \centering
    \scalebox{0.85}{
    \begin{tabular}{cccccccc}
    \toprule
    \textbf{Datasets} & \textbf{Years}& \textbf{Domain(s)}  & \textbf{Language(s)}   & \textbf{No of docs} & \textbf{No of words} & \textbf{No of terms} &\textbf{No of entities}\\  \midrule
    TTC\footnote{\url{https://www.clarin.si/repository/xmlui/handle/11356/1463/}}  & 2012 &  \begin{tabular}[c]{c@{}c@{}} Wind energy,  \end{tabular} & Chinese &270&238,299&293&-\\ \cline{4-8}
    &  &  Mobile technology &English    &209&51,699&269&-\\\cline{4-8}
    &  & &French     &-&-&-&-\\ \cline{4-8}
    &  & &German     &-&-&-&-\\ \cline{4-8}
    &  & &Latvian    &-&-&-&-\\ \cline{4-8}
    &  & &Russian    &-&-&-&-\\ \cline{4-8}
    &  & &Spanish    &-&-&-&-\\ \midrule
    BitterCorpus\footnote{\url{https://mt4cat.fbk.eu/benchmarks/bittercorpus}} & 2014& Information Technology  & English  & 56 & - &874&-\\ \cline{4-8}
    & & & Italian   &56&-&874&- \\ \midrule
    TermFrame v1.0\footnote{\url{http://termframe.ff.uni-lj.si/}} & 2019&Karstology & Slovene  &60&987,801 &2,946&341\\ \cline{4-8}
    & & & Croatian   &43&969,735&-&-\\ \cline{4-8}
    & & & English    &54&1,968,509 &4,397&156 \\ \midrule
    KAS-biterm\footnote{\url{https://www.clarin.si/repository/xmlui/handle/11356/1263}} & 2019 & Academic writing &Slovene &176&-&1,895&-\\ \cline{4-8}   
    & & & English &176&-&2,138&-\\ \cline{4-8}
    & & & others  &176&-&168&-\\ \midrule
    ACTER v1.5\footnote{\url{https://github.com/AylaRT/ACTER}} & 2020 &  Corruption, Dressage, & English & 248 & 222,281 & 5,534 &1,333\\\cline{4-8}
    &  & Heart failure, &French    &309&240,207&4,941&789\\ \cline{4-8}
    &  & Wind energy&Dutch    &259&218,989&5,454&884\\ 
    \bottomrule
    \end{tabular}
    }
    \label{tab:multi-dataset}
\end{table}

Also, other corpora focused on specific domains in multiple languages. The BitterCorpus \cite{arcan2014enhancing} was released in 2014 as a manually-annotated collection of parallel English-Italian documents with 874 domain-specific bilingual terms in the Information Technology (IT) domain. The documents are extracted from two corpora: the GNOME and the KDE data collections. The summary for these two corpora and the BitterCorpus itself can be found on this website\footnote{\url{https://mt4cat.fbk.eu/benchmarks/bittercorpus}}.

TermFrame v1.0 \cite{vrtovec2019termframe,pollak2019karst} is a specialized corpus of karstology literature in Slovene, Croatian, and English. The terms were extracted from 24 English and 60 Slovene documents by comparing the domain corpus to the reference corpus, and with additional processing steps (e.g., filtering based on nested terms, stopwords, and fuzzy matching). The summary for each domain can be found on this website\footnote{\url{http://termframe.ff.uni-lj.si/}}.

KAS-biterm \cite{ljubevsic2019kas} is a collection of 700 bilingual Ph.D. theses (40 million tokens) in the KAS corpus about Slovene academic writing of three domains: Chemistry, Computer Science, and Political Science. The sentences in the corpus were filtered by noSketch Engine to keep only those with a high chance of containing the term in the original language and its translation into Slovene. The terms were labeled using three labels (terms, partial terms, abbreviations) in three categories of languages (Slovene, English, or other languages). The summary for each domain can be found on this website\footnote{\url{https://www.clarin.si/repository/xmlui/handle/11356/1199}}.

Annotated Corpora for Term Extraction Research (ACTER) dataset \cite{rigouts2020termeval} was the first meticulously annotated corpus that covered multiple languages and domains and that offered available up-to-date documentation and transparent annotation guidelines. Texts from four different domains (Corruption, Dressage, Heart failure, and Wind energy) and three languages (English, French, and Dutch) were manually annotated. The texts for a specific domain are comparable across languages, meaning that texts from the same domain but across different languages cover the same subject. While most texts cannot be aligned as parallel translations, the texts from the corruption domain are sentence-aligned. Furthermore, ACTER solves the debate about whether or not to consider named entities as terms by providing two different versions of manual annotations: one containing only terms, and the other containing both terms and named entities. As of yet, these characteristics make ACTER the most useful resource for benchmarking and method comparison in both monolingual and multilingual ATE. The summary for each domain can be found on this website\footnote{\url{https://github.com/AylaRT/ACTER}}.

While we investigated different available corpora, we figured out that most of the original papers fully describe the primary version of the dataset. Some papers included both data and method descriptions but rarely documented data versions of the dataset and whether an entire dataset was used or only a part of it. As a result, given the same corpora, it is hard to make a thorough comparison among methods given they use different data versions, especially when the latest version has a significant update compared to the previous one. 

\section{Supervised ATE systems}
\label{sec:systems}

In this section, we offer an extensive overview of recent ATE systems, which tend to rely either on shallow machine learning techniques or deep neural networks. In most cases, traditional approaches to term extraction do not rely on machine learning. Instead, they tend to extract candidate terms above a certain threshold, derived from several linguistic and statistical features. Traditional approaches are not covered in this survey, since they have been covered in previous surveys (e.g., \cite{kageura1996methods} and \cite{astrakhantsev2015methods}). Instead, in this section, starting from the advent of machine learning up until now, we present machine learning-based (neural and non-neural) ATE systems, that have not been systematically described and compared in any recent survey study.

\subsection{Machine learning approaches}

\begin{figure}[ht]
    \centering
    \includegraphics[scale=0.4]{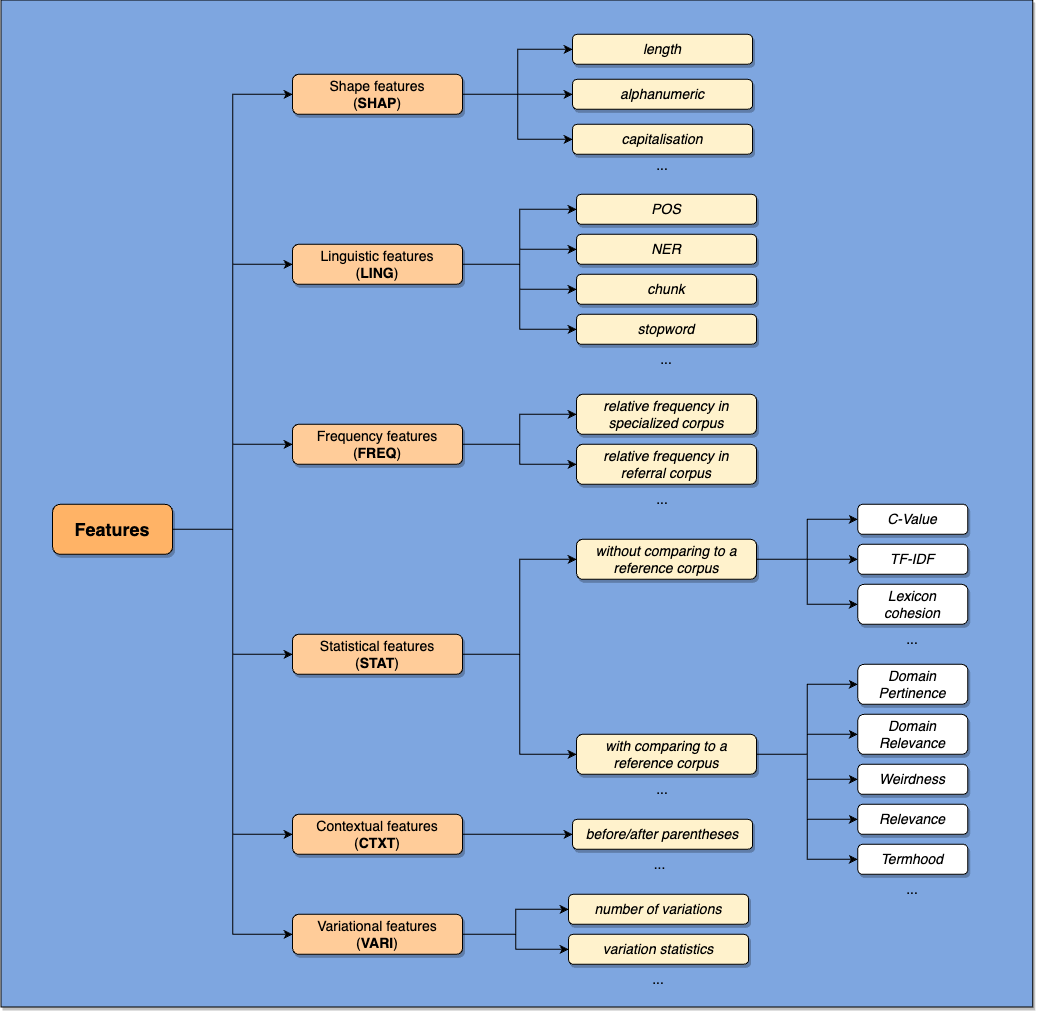}
    \caption{Feature group and subgroup for machine learning models based on \cite{rigouts2021hamlet}'s work.}
    \label{fig:features}
\end{figure}

Despite several variations in features and models, most supervised machine learning approaches to ATE comply with the traditional approach, following three principal steps: (1) preprocessing, (2) feature engineering, and (3) term extracting classifier. In the preprocessing step, some operations are performed on input texts (e.g., sentence segmentation, word segmentation, and POS tagging) to make the input text ready for further steps. In the feature engineering step, we describe the candidate terms through several different features (see examples of different types of features in Figure \ref{fig:features}). In the final steps, the features are fed into the machine learning classifiers so that the classifiers can learn from the training set and use that knowledge to make predictions on new unseen texts.

Due to the relatively low accuracy, these first machine-learning approaches were mostly used to complement approaches based on hand-crafted rules. This, along with the success of traditional systems (e.g., TermoStat \cite{drouin2003term}), which relied on several linguistic and statistical features, lead to the idea of combining different types of information. That is to say, multiple linguistic and statistical termhood indicators were fed as features to a variety of machine learning algorithms. While several NLP tools (e.g., tokenization, lemmatization, stemming, chunking, POS tagging, and full syntactic parsing) are employed in this approach to obtain linguistic profiles of term candidates, numerous statistical measures are also applied to this approach, including the termhood \cite{vintar2010bilingual}, unithood \cite{daille1994towards}, C-value  \cite{frantzi1998c}. Regarding machine learning algorithms, the most popular algorithms used for ATE include AdaBoost \cite{castellvi2001automatic}, ROGER evolutionary algorithm \cite{aze2005preference},  RIPPER rule induction \cite{foo2010using}, CRF++ \cite{judea2014unsupervised}, K-nearest neighbors \cite{qasemizadeh2014evaluation}, Logistic Regression \cite{dobrov2011multiple,loukachevitch2012automatic,nokel2012combining,bolshakova2013topic,fedorenko2014automatic}, Decision Trees \cite{karan2012evaluation}, and Support Vector Machines \cite{ljubevsic2018kas}. 

An example of a machine learning approach employing extensive feature engineering and several classifiers is given in \cite{conrado-etal-2013-machine}. In this study, the authors proposed to select statistical and linguistic features and feed them into different machine learning classifiers (e.g., JRip, Naive Bayes, J48, or SMO from WEKA). Several post-processing steps were carried out (e.g., word variation standardization, POS annotation, and stopword and punctuation removal) before the selection of the final candidate terms. 

\cite{yuan2017supervised} instead proposed a machine learning method with common features (e.g., term frequency, c-value, weirdness) extracted from the token n-grams (n={1,2,3,4,5}) excluding all the stopwords. The selected features were fed into several different classifiers, namely Random Forest, Linear Support Vector Machine, Multinomial Naive Bayes classifier, Logistic Regression, and SGD classifiers. 

With the advent of the newly annotated ACTER corpora, one of the TALN-LS2N approaches \cite{hazem2020termeval} used the combination of different meaningful information—such as linguistic, stylistic, statistic, and distributional descriptors—to generate the feature vectors. It then made use of the XGBoost model to learn several classifiers, which were weighted according to their performance and aggregated iteratively.

The HAMLET approach \cite{rigouts2021hamlet} included a relatively wide range of supervised machine learning methods (e.g., Decision Tree, Random Forest, Multi-layer Perceptron, and Logistic Regression) and relied on a total of 152 features from six different feature groups: morphological, frequency-based, statistical, relational and linguistic, and corpus-based.
The Binary Random Forest classifier performed the best out of all tested classifiers.

Furthermore, \cite{nugumanova2022nmf} proposed a new approach that combines probabilistic topic modeling (PTM) and non-negative matrix factorization (NMF). They compared five different NMF algorithms and four different NMF initializations, and found optimal combinations of NMF to compare with the extraction baseline (e.g., TF-IDF, RAKE, YAKE, and TextRank).

\subsection{Deep learning approaches}

The use of neural networks, especially language models, to solve ATE, has been gaining a lot of traction recently. Their application is performed either to just represent the information in the text with word embeddings or to apply a deep architecture as an end-to-end classifier. In this section, we describe studies that employed any of these two approaches.

\subsubsection{Embeddings}

The embeddings for term extraction are often pretrained on a large general corpus, and then potentially fine-tuned on specific corpora during classification. The most popular general non-contextual word embeddings that were used in several ATE studies \cite{amjadian2018distributed,kucza2018term,zhang2018adapted,le2021multilingual} are GloVe embeddings\footnote{\url{https://nlp.stanford.edu/projects/glove/}} \cite{pennington2014glove}. The second most popular option is constituted by domain-specific Word2Vec embeddings \cite{mikolov2013linguistic}, either employing the CBOW or the skip-gram architecture \cite{amjadian2018distributed,wang2016featureless,bay2021term} and FastText \cite{terryn2022tagging}.  

The first study to successfully apply word embeddings for the ATE task was the one by \cite{amjadian2016local}. The embeddings, pretrained on a general corpus, were initially used in a filtering step on top of an existing rule-based tool TermoStat \cite{drouin2003term}, and later as a standalone term extraction tool \cite{amjadian2018distributed}. Meanwhile, SemRe-Rank \cite{zhang2018semre} made use of Word2vec embeddings to create semantic relatedness graphs of words, thus computing their ``semantic importance'' scores. These scores were then used to revise the base scores of term candidates computed by another ATE algorithm.

Several papers also researched domain-specific embedding representations for ATE usage in a specific field, for example, as \cite{bay2021term}'s research in medicine. Meanwhile, some studies attempted to combine  domain-specific and general embeddings. \cite{amjadian2018distributed} explored the idea of concatenating general and domain-specific embeddings as an input (in English) for a Multi-Layer Perceptron (MLP) from the 1M+ token mathematics corpus. Similarly, \cite{hatty2020predicting} tested two new (neural) approaches to exploit general vs. domain-specific comparisons. They used a simple neural network model with pre-computed Ge-embedding information as input, and a multi-channel model computing the comparison internally. These embedding concatenation techniques demonstrated that any strategy using both types of embeddings performed better than the strategies only using either the general or the domain-specific ones.

Considering the importance of general knowledge encoded in the pretrained language model and contextual information for the downstream task, the employment of several contextual embeddings for the ATE task was proposed. This is the case, for example, of the (stack) bi-directional Flair embeddings \cite{terryn2022tagging}  with a neural, character-based language model from the FlairNLP framework\footnote{\url{https://github.com/flairNLP/flair}} for multiple languages that incorporate subword information, and newly adopted Transformer-based (monolingual and multilingual) embeddings, some of which were already tested in term extraction (e.g. BERT embeddings \cite{andrius2020automatic,terryn2022tagging} and its variants), or the stacked Flair + BERT embeddings \cite{terryn2022tagging}.


\subsubsection{ATE Classifiers}

\begin{figure}[ht]
    \centering
    \includegraphics[scale=0.48]{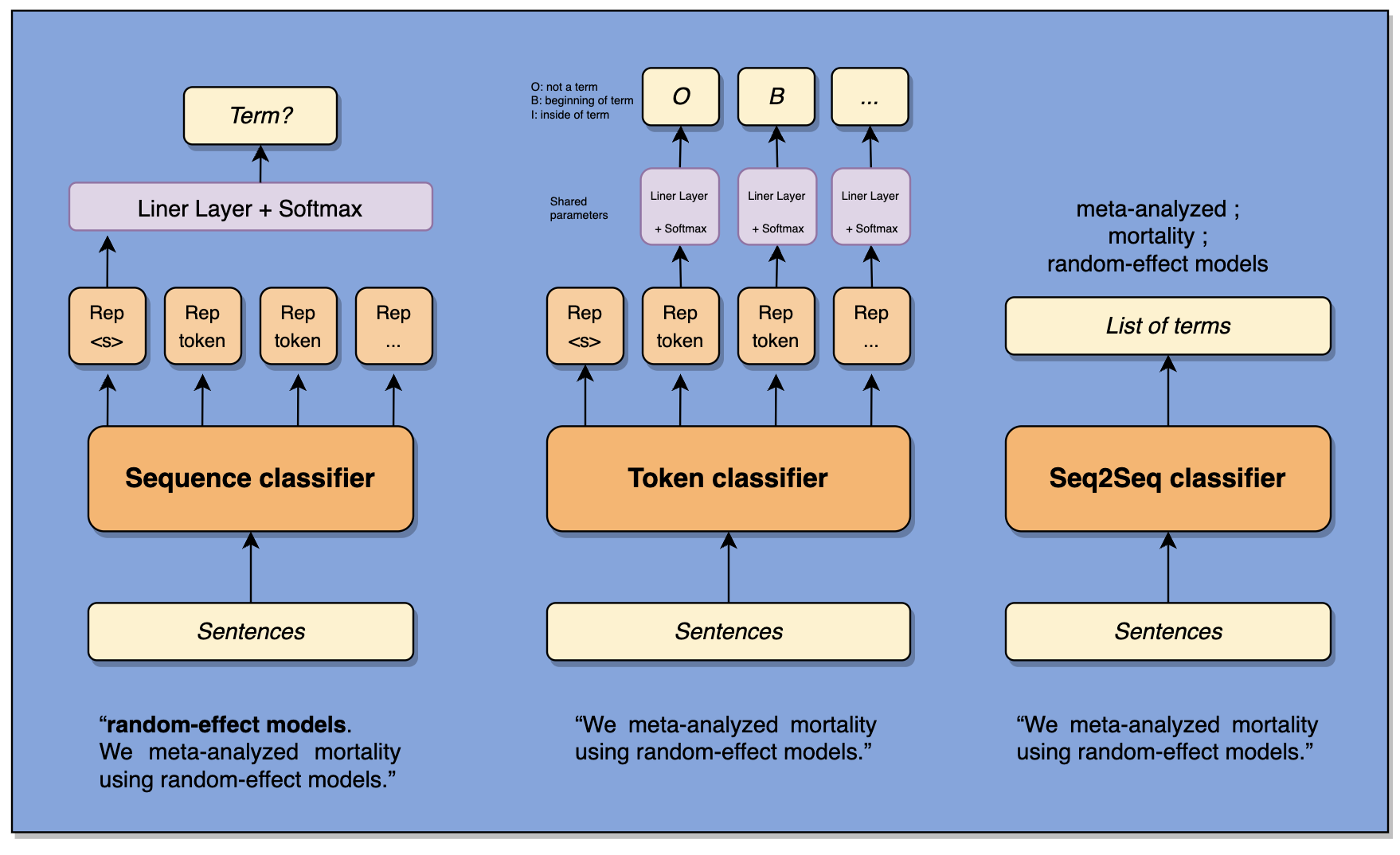}
    \caption{An example of how three types of neural ATE classifiers work (from left to right: Sequence classifier (e.g., XLMR); Token classifier (e.g., BERT); Seq-2-seq (e.g., mBART).}
    \label{fig:data_structure}
\end{figure}

The use of deep neural networks in ATE is not only limited to the generation of embedding representations. Neural architectures are also used as end-to-end term extractors. Based on the methodology, we can divide neural extractors into three main types, as demonstrated in Figure \ref{fig:data_structure}: (1) sequence classifiers; (2) token classifiers; and (3) sequence-to-sequence generation models, which are used for several generation tasks (e.g., summarization, translation, and headline generation).

When deep neural models were first adopted for ATE, ATE was considered as a sequence or binary classification task. The resulting approaches relied on the positive (is-a-term) and negative (not-a-term) samples, which are generated from all possible n-grams of a fixed length of a given sentence. TALN-LS2N \cite{hazem2020termeval} was the first team to propose this, using BERT-based language models in the recent ATE challenge TermEval 2020. Their approach outperformed all classification methods based on feature engineering. They tested different versions of BERT (RoBERTa for the English corpus, CamemBERT for the French corpus) as binary classifiers for term prediction. The input consisted of a sentence and a selected n-gram within the sentence. If the n-gram was a term, the input was labeled as a positive training example, otherwise as a negative example. Using the same approach and training datasets, \cite{lang2021transforming} improved the performance of the ATE binary classifier by employing a multilingual version of RoBERTa, the so-called XLM-RoBERTa (or XLM-R, for short) and compared it with other neural approaches. Although this approach proved to have better predictive power in comparison with other machine-learning-based approaches, generating all possible n-grams contained in each sentence across all documents in order to create the training set is computationally expansive and space-demanding. For this reason, later studies decided to shift from this approach toward sequence labeling or token classification.

The latest trend is the use of language models and sequential methods where the candidate terms are detected in their original contexts, usually by classifying each token in the text sequence as (part of) a term or not. Thus, instead of having a label for the whole text sequence, the token classifier assigns a label for each token. This sequence labeling approach was first tested by \cite{judea2014unsupervised}, who proposed a non-neural conditional random field (CRF) classifier to investigate the utility of different types of features. \cite{wang2016featureless} 
introduced co-training—a weakly supervised bootstrapping paradigm that helps in stabilizing classifier's predictions—using a long short-term memory (LSTM) network and a convolutional neural network (CNN). Alternatively, \cite{kucza2018term} applied a token classifier by taking advantage of recurrent neural networks (RNN). \cite{han2018cnn} proposed the CNN-BiLSTM-CRF model to minimize the influence of different word segmentation results on Chinese term extraction. Another attempt at deep learning based sequential labeling for ATE task is presented in \cite{andrius2020automatic}. This study  experimented on several models, such as LSTM, BiLSTM, LSTM-CRF, and BiLSTM-CRF, with the FastText and BERT embeddings, to obtain a high evaluation score for ATE in the cybersecurity domain from the less-resourced Lithuanian language. 

Regarding more recent competitions, NLPLab UQAM \cite{le2021multilingual}, the team which won in the use of TermEval's Dutch corpus, also modeled the task as token classification by applying a bidirectional LSTM (biLSTM) with pretrained GloVe word embeddings. The monolingual version of D-Terminer \cite{rigouts2022d} and \cite{hazem2022cross} took advantage of the Flair framework to propose a feature-based strategy that consists in using BERT embeddings as input features for a biLSTM with a CRF layer. Given the same dataset, \cite{terryn2022tagging} made a comparison between the neural sequence-labeling approach and the traditional feature-based approach, where they proposed the evaluation of three-word embeddings (i.e., FastText, Flair, and BERT) in monolingual and multilingual settings. 

With the advent of Transformer-based models, several pretrained language models have been applied as token classifiers to boost the extraction performance \cite{hazem2022cross,lang2021transforming}. Above all, XLM-R \cite{conneau2019unsupervised} was applied and mentioned in several publications and is now considered a benchmark for several languages \cite{trantransformer}. Since this approach operates without upfront n-gram generation and handles each sentence as a single example that needs to be labeled, it is considerably more time efficient than the previous sequence classification approach. 

Despite the widespread use of Transformer-based sequence-to-sequence generation models for NLP, the adaption of self-supervised pre-training approaches for ATE has just recently begun to gain traction. \cite{lang2021transforming} was the first to employ the sequence generation model mBART \cite{liu2020multilingual}) to transform the input sentences to sequences of comma-separated terms. This approach was inspired by the neural machine translation-based ontology learning proposed by \cite{petrucci2018expressive}. While promising, the performance and applicability of this approach require additional testing and remain an unsolved issue.

\subsection{Multilingual learning}

Recently, a new research direction in ATE aims to conduct term extraction in a multilingual setting, testing the hypothesis that multilingual information can help during the monolingual extraction phase.

Multilingual term extraction can be performed on parallel or comparable corpora. According to \cite{mcenery2006corpus}'s definition, a comparable corpus is composed of monolingual data collected from different languages using the same sampling techniques. Meanwhile, a parallel corpus is one that is composed of source texts and their translations in one or more different languages. While the comparable corpora are not translations, they do share some vocabulary (e.g., cognates) due to the fact that they cover the same topic. This can be exploited to extract equivalent terms found in parallel corpora, usually by leveraging term alignment approaches.

Regarding parallel corpora, \cite{foo2012computational} proposed to divide the methods into two groups: ``align-extract'' and ``extract-align'', depending on whether monolingual candidate terms are extracted first, or whether the alignment is performed first. Several tools have been proposed (e.g., the TExSIS tool \cite{macken2013texsis}) for bilingual automatic term extraction from parallel corpora. The TExSIS tool extracts word alignments with GIZA++ \cite{och2003systematic} and applies the rule-based chunking \cite{macken2010chunk} with alignment progress and additional statistical (termhood) filtering to extract the candidate terms. Besides GIZA++, some of the most popular methods for the alignment \cite{ivanovic2022corpus} are Moses phrases tables \cite{koehn2007moses}, Awesome Align neural word alignment \cite{dou2021word}, and ASTrED aligned syntactic tree edit distance \cite{vanroy2021metrics}, to mention a few.

Despite the significant efforts that have been made to create multilingual corpora that cover more than one domain, parallel multilingual datasets have been rarely created for less-resourced specialized domains. This is due to the fact that the creation of these datasets tends to be error-prone, labor-intensive, and time-expensive, and was also criticized in the past for subjectiveness. Instead, most of the multilingual corpora are comparable, and multilingual language models are commonly used for term extraction on these corpora. An example of this kind of approach is HAMLET \cite{rigouts2021hamlet}. \cite{rigouts2021hamlet} proposed six different settings, in which they used 5-fold cross-validation to train and test on a single corpus (one domain, one language), single language (all domains, one language), or all corpora (all domains, all languages), with or without POS patterns. \cite{trantransformer} leveraged the multilingual setup by fine-tuning the model using training datasets from all the languages in the ACTER corpora (English, French, and Dutch) and the Slovenian RSDO5 corpus, and then applying the model to the test sets of all languages. By doing so, they examined whether adding more data from other languages to the training set that matches the target language in the testing set improves the predictive performance of the model.

\subsection{Cross-domain and cross-lingual transfer learning}

Current research on term extraction deals with two major issues: (1) the data collection for a specific field of expertise or domain is often non-existent or small, and (2) the available well-annotated corpora are scarce, especially for lesser-known languages (e.g., Slavic languages). To address the former issue, \cite{liu2018towards} proposed to combine specialized and general domain word embeddings to increase the training corpora. However, this technique is hard to apply for most ATE tasks, since it still requires a ``reasonable'' amount of domain-specific manually annotated data, which is scarce (i.e., the second issue). To solve both problems, cross-domain and cross-lingual approaches have been recently introduced.

\subsubsection{Cross-domain learning}

Zero-shot and few-shot learning recently gained in popularity as advanced approaches to solve the scarcity of available annotated corpora. They rely on transferring knowledge the model obtains from a well-resourced domain/language with plenty of annotated data to a less-resourced domain/language with scarce or no data. This way, cross-domain learning can also be used as a way of checking the generalization capabilities (and therefore usefulness) of the model, i.e., how successfully the knowledge that the model obtains on one domain can be applied to the arbitrary new unseen domains.

\cite{hazem2022cross} was the first to propose to fine-tune several Transformer-based language models and to test them in a cross-domain scenario. In all the scenarios, two domains were used for training, one for validation, and one for testing. The authors proposed two evaluation settings: 1) fixing the training data sets while varying BERT models to test different model versions, and 2) fixing the models while varying the training data sets to test the success of different cross-domain transfers. The results suggested that while cross-domain learning for ATE is possible, the models are still domain sensitive, and a careful choice of training datasets is necessary to achieve competitive performance.

\cite{lang2021transforming} used the same cross-domain setting with the same corpora, and experimented with three different Transformer-based term extraction models operating on sentence level: a language model for token classification; one for sequence classification; and innovative use of sequence to sequence (seq2seq), which learns to reduce sentences to terms. \cite{trantransformer} tested the same cross-domain setting on a less-resourced language, Slovenian (RSDO5 corpus), using a multilingual model. They showcased a generalization of cross-domain transfer learning also on Slovenian. Finally, \cite{tranicadl2022} employed several Transformer-based multilingual and monolingual model ensembles in a cross-domain setting on both the ACTER and RSDO5 dataset, further improving the results of both previous studies and confirming the success of the cross-domain transfer.

\subsubsection{Cross-lingual learning}

Similarly to cross-domain learning, cross-lingual learning is based on transferring knowledge obtained by the model from annotated data in a high-resourced language to the input data in a less-resourced language. We examine how well the ATE model performs without the language-specific training corpus and how good the knowledge transfer between different languages is. 

\cite{hazem2022cross} tested several multilingual fine-tuned Transformer language models in a zero-shot cross-lingual setting. They did so by fine-tuning the model on the ATE task in one or several languages, and by testing them on a new target language, on which the model was not fine-tuned. The results indicate a successful transfer,  which led to achieving competitive results on the ACTER dataset and opening a new promising direction in ATE. \cite{lang2021transforming,tranicadl2022} further extended this work with a new benchmark language model, namely XLMR \cite{conneau2019unsupervised}, testing its performance in a cross-lingual setting and using the sequence labeling approach. The results re-confirmed the potential of transferring knowledge from rich-resourced to less-resourced languages in the field of ATE since this technique allows us to beat the previous state-of-the-art methods on the ACTER dataset. 

\section{ATE evaluation metrics}
\label{sec:eval}

\begin{figure}[ht]
    \centering
    \includegraphics[scale=0.65]{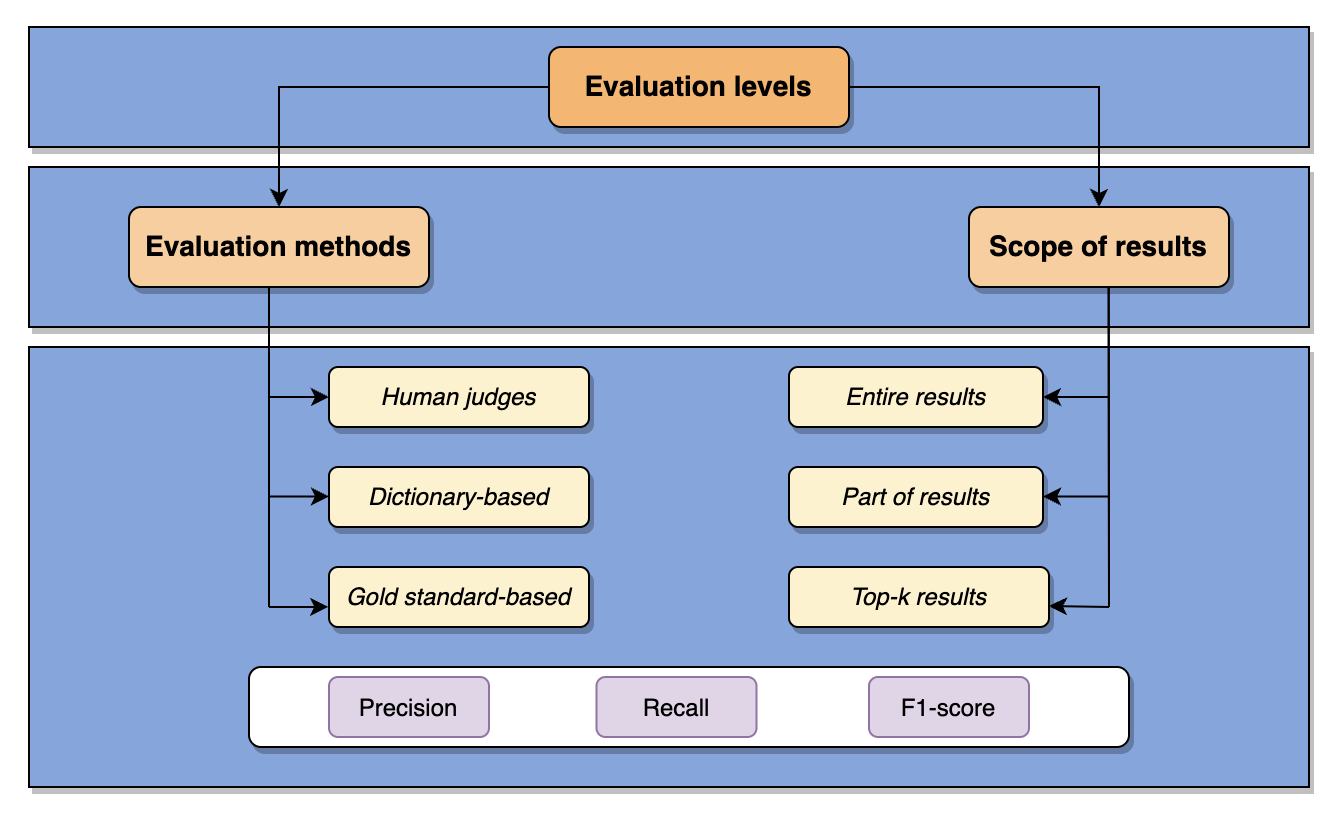}
    \caption{The overview of different evaluation metrics in ATE task.}
    \label{fig:evaluation}
\end{figure}

ATE models usually provide a list of candidate terms from the given domain-specific corpus as the final output, and it is important to define the correct methods and metrics to evaluate the quality of this output. There are several variations among the evaluation methods, including both intrinsic and extrinsic mechanisms. The extrinsic methods assess the quality of the extracting system by measuring the improvement in the performance of another system or application that uses the results of term extraction as described in \cite{vivaldi2007evaluation}. Meanwhile, the intrinsic ones measure the quality of term extractor by evaluating some intrinsic properties, which is independent of their intended planned use. We can classify the direct evaluation approaches into two main aspects: the evaluation methodology and the scope of the results \cite{zhang2008comparative} as shown in Figure \ref{fig:evaluation}. While the evaluation methodology focuses on whether the evaluation is performed by human judges, or whether it is a dictionary-based or gold standard based, the scope of results answers the question of whether we evaluate the entire results, parts of the results, or the top-k results. Due to this variety in evaluation types, the performance of different approaches is often not directly comparable.

Regarding the evaluation methods, in the initial studies about ATE, having a human judge was one of the first approaches to evaluate how well the candidate terms were extracted. \cite{justeson1995technical} evaluated their system by asking a domain expert (i.e., a terminologist) to judge whether the extracted candidates were domain-specific terms.

In order to reduce the human effort, the dictionary-based evaluation was proposed either in the form of a list or as a dictionary of the pre-defined criteria to map with the extracted terms. \cite{kageura1996methods} took advantage of statistical features, e.g., termhood and unithood, for evaluation. \cite{l1996definition} applied five pre-evaluation criteria from the basic design evaluation to present how the results are shown in the candidate lists. \cite{macken2013texsis} identified the terms with linguistic preprocessing steps and matched the candidate terms to a pre-defined dictionary of POS patterns. 

Another popular approach is based on reference term lists, or the so-called gold standard corpora, which can be an adaptation of a pre-existing list, a sample (seed terms), or the list of all the terms in the corpus. The pre-existing term list \cite{wermter2005massive,dobrov2011multiple} is not collected directly from the training corpora but is an already existing and community-wide terminology. Thus, the term extraction approaches using the corpora with this gold standard often evaluate how many candidate terms were actual terms but failed to measure how many terms in the text were correctly extracted. Meanwhile, the approach of considering a sample as the gold standard \cite{baroni2004bootcat,loginova2012reference} took advantage of a web crawler to collect the domain-specific texts (e.g., TTC project). The crawler took a list of domain-specific words, the so-called seed terms, as input, and as outputs, the texts found on the Web in the domain of interest. The seed terms are usually term representative of the domain for which we want to retrieve the web documents. To resolve the existing issues of the two mentioned types, the list of all the terms in the corpus \cite{kim2003genia,rigouts2020termeval} was used as a gold standard where the list was annotated directly from the corpora. This became a benchmark for manual annotation in term extraction until now. If this gold standard corpus is used, the final evaluation score is usually obtained by calculating Precision (i.e., how many candidate terms were actual terms), Recall (i.e., how many terms in the text were correctly extracted), and F1-score (i.e., the harmony mean of Precision and Recall) for the obtained list of term candidates. We will specify these metrics when discussing the evaluation from different scopes of results.

Finally, besides the performance of the ATE system in terms of the evaluation metrics described above, some claim that other factors, such as the consistency of the predicted candidate terms, are also important. For example, \cite{sauron2002tearing} proposed that the measures of a model's quality should also concern reliability, efficiency, maintainability, usability, and portability.

Regarding the scope of results, the most common ATE evaluation approach is to compare the candidate term lists obtained by the system against the list of terms extracted by human annotators (i.e., the so-called gold standard corpus \cite{kim2003genia,conneau2019unsupervised,rigouts2020termeval,rigouts2021hamlet,tranicadl2022}). This is done by calculating Precision (i.e., how many candidate terms are actually terms, see Equation \ref{eq:eval1}), Recall (i.e., how many terms in the text were correctly extracted, see Equation \ref{eq:eval2}), and F1-score (i.e., the harmonic mean between Precision and Recall, see Equation \ref{eq:eval3}). 

\begin{equation}
\label{eq:eval1}
\small
Precision = \frac{Number \: of \: correctly \: terms}{All \:  the \: candidate \: terms} = \frac{True \: Positives}{True \:  Positives + False \: Positive}
\end{equation}
\begin{equation}
\label{eq:eval2}
\small
Recall = \frac{Number \: of \: correctly \: terms}{All \: terms\: in \: the \: corpus}  = \frac{True \: Positives}{True \:  Positives + False \: Negative}
\end{equation}
\begin{equation}
\label{eq:eval3}
\small
F1-score = 2 \cdot \frac{Precision \cdot Recall}{Precision + Recall}
\end{equation}

Several ATE systems use the same evaluation approach as the Keyword Extraction task, i.e., they choose top-k best candidate terms for evaluation (top-100 candidate terms \cite{kupiec1993algorithm,ideue2011comparison}, top-300 candidate terms \cite{vintar2010bilingual}, top-500 candidate terms \cite{daille1994study}). \cite{macken2013texsis} and \cite{zhang2018adapted} evaluated the best-k candidate terms by using a variable k. Most of these studies employed Precision for evaluation of the best-k terms \cite{drouin2003term,sclano2007termextractor,macken2013texsis,zhang2018adapted,haque2018termfinder}. 

One notable insight which emerged from our literature review is that the number of studies using just Precision is higher than the number of studies that evaluate the system according to all three evaluation criteria (i.e., Precision, Recall, and F1-score). This is due to the relative lack of available gold standard data in the past. However, thanks to the more recent effort of constructing and manually annotating domain-specific datasets for ATE, most of the current systems employed on contemporary benchmark datasets (e.g., ACTER) are evaluated according to all three metrics \cite{rigouts2020termeval,rigouts2021hamlet}.

\section{Comparative results}
\label{sec:results}

Since term extraction methods vary greatly in regard to definition, corpora, domains, languages, and evaluation metrics, a comparative evaluation of term extraction methods is hardly achievable. Nevertheless, in this section, we propose to measure the performance of several ATE systems by comparing the candidate term list extracted on the whole test set level with the manually annotated gold standard of each domain using a strictly matching F1-score.  All approaches are applied to the newest and the most popular up-to-date ATE corpus, namely the ACTER dataset. We chose this dataset for evaluation because it is the most systematically annotated corpus that covers multiple languages and domains, and that also contains available up-to-date documentation and transparent annotation guidelines. The results of the evaluation are presented in Table \ref{res:acter}, including both the approaches evaluated on the ACTER dataset in the related work (non-neural models and some deep learning models) and our own experiments on the same dataset.

\begin{table}
\caption{The F1-score evaluation of benchmark approaches on the Heart Failure test set from the ACTER corpus in three languages (English (EN), French (FR), and Dutch (NL)) and two types of annotation, with named entities (NER) and without named entities (ANN). The best approach for each category is highlighted in bold.}
\label{res:acter}
\centering
\begin{tabular}{ccccccc} 
\toprule
\textbf{Methods}           & \multicolumn{6}{c}{\textbf{ACTER}}    \\

                          & \multicolumn{3}{c}{\textbf{ANN}} &  \multicolumn{3}{c}{\textbf{NES}}\\
\toprule
\rowcolor{blue}
\textbf{Non-neural models}  & EN & FR & NL    & EN & FR & NL       \\ 
\midrule
RACAI \cite{rigouts2020termeval}                            &    39.3     &     -   & - &   41.3    & -       &    -        \\
e-Terminology \cite{rigouts2020termeval} &  21.4       &   20.6     &     15.3  & 20.1  &    19.7    &    14.4        \\
NYU \cite{rigouts2020termeval}    &    31.5     &    -    &  - &   30.6      &  -      &     -      \\
HAMLET \cite{rigouts2021hamlet}  & \textbf{54.2} &    \textbf{60.2}     &   \textbf{66.1}&       \textbf{55.4}      &    \textbf{60.8}    &  \textbf{66.0}       \\
Feature-based XGBoost \cite{hazem2022cross} &   -       &   -     &  -  &    33.6     &   50.9     &    34.1      \\ 

NMF \cite{nugumanova2022nmf}  & 33.5 & 30.9 & 30.1 & 33.7 & 30.7 & 30.3 \\
\toprule
\rowcolor{blue}
\multicolumn{7}{l}{\textbf{Monolingual deep learning models}}      \\ 
\midrule
TALN-LS2N \cite{hazem2020termeval}  &   45.0       &   45.9     &  -   &   46.7      &  48.1      &     -    \\
NLPLab UQAM \cite{le2021multilingual}   &   17.8       &   12.9     & 18.6   &    18.1     &    13.2    &    18.7      \\ 
XLM-R sequence Classifier \cite{lang2021transforming}&-&-&-&45.2&46.0& 48.5         \\
XLM-R Token Classifier \cite{lang2021transforming}&-&-&-&58.3&52.9 &  \textbf{69.6}        \\
mBART NMT \cite{lang2021transforming}&-&-&-& 53.2& 55.9 &   65.2       \\
Vanilla-biLSTM-CRF  \cite{hazem2022cross}&-&-&-&8.17& 6.53&7.50\\
BERT   \cite{hazem2022cross}&-&-&-&48.2&-&-\\
CamemBERT   \cite{hazem2022cross}&-&-&-&-&51.1&-\\
BERT (NER)  \cite{hazem2022cross}&-&-&-&37.4&-&51.0\\
CamemBERT (NER)  \cite{hazem2022cross}&-&-&-&-&51.1&-\\
BERT-biLSTM-CRF   \cite{hazem2022cross}&-&-&-&29.5&25.6& 27.4\\
mBERT  \cite{hazem2022cross}&-&-&-&45.8&47.2&-\\
ALBERT  \cite{tran2022ensembling} &49.9&-&-&54.5&- &-\\
DistilBERT  \cite{tran2022ensembling} &46.5&-&-&53.9&- &-\\
ELECTRA  \cite{tran2022ensembling} &51.2&-&-& 52.2&- &-\\
RoBERTa  \cite{tran2022ensembling} &54.3&-&-&59.1&- &-\\
XLNet  \cite{tran2022ensembling} &55.2&-&-& \textbf{57.8}&- &-\\
FlauBERTa  \cite{tran2022ensembling} &-&38.9&-& -&51.4 &-\\
Dutch BERT  \cite{tran2022ensembling}&-&-&65.6& -&- &66.8\\
robBERTa  \cite{tran2022ensembling} &-&-&48.2& -&- &62.2\\
robBERTa-v2  \cite{tran2022ensembling} &-&-&48.3& -&- &63.4\\
mInfoXLM \cite{tran2022ensembling} &\textbf{56.1}&50.4&\textbf{68.5}&57.6&58.0&68.6\\
mBERT \cite{tran2022ensembling} &43.0&48.4&66.3&55.0&\textbf{57.4}&67.8\\
mDistilBERT \cite{tran2022ensembling} &49.8&\textbf{51.9}&65.3& 55.6& 55.8&67.8\\
\toprule
\rowcolor{blue}
\multicolumn{7}{l}{\textbf{Multilingual deep learning models}}      \\ 
\midrule
XLM-R Sequence Classifier \cite{lang2021transforming}&-&-&-&46.0& 46.7& 56.0         \\
XLM-R Token Classifier \cite{lang2021transforming}&-&-&-&\textbf{56.2}& 55.3 &  \textbf{67.8}        \\
mBART NMT \cite{lang2021transforming}&-&-&-& 55.3&\textbf{57.6} &   64.9      \\
mBERT  \cite{hazem2022cross}&-&-&-&45.4&44.9&51.0\\
\toprule
\rowcolor{blue}
\multicolumn{7}{l}{\textbf{Cross-lingual deep learning models}}      \\ 
\midrule
XLM-R Sequence Classifier \cite{lang2021transforming}&-&-&-&44.7  & 48.1 &  58.0      \\
XLM-R Token Classifier \cite{lang2021transforming}&-&-&-&\textbf{58.3}  & 57.6 & \textbf{69.8}       \\
mBART NMT \cite{lang2021transforming}&-&-&-&  55.2 &  57.4 &   59.6      \\
XLM-R Token Classifier \cite{tran2022can} &   \textbf{56.0}      & \textbf{59.9} & \textbf{68.5}       &58.0    & \textbf{61.1}        &     68.3     \\ 
\bottomrule
\end{tabular}
\end{table}

Note that deep learning-based approaches are divided into three different categories based on how they took advantage of languages in the training corpus. While monolingual deep learning models use a single language for training and predict on the new unseen data from the same language, multilingual deep learning models are fine-tuned on multiple languages at once (e.g., on a combination of English, French, and Dutch data in the ACTER corpus) and generate a prediction on the new unseen data from all the languages. Finally, cross-lingual deep learning models are trained on one (or more) languages and generate predictions on unseen data from a new language that the model did not encounter during the training. 

TALN-LS2N, RACAI, NYU, e-Terminology, and NLPLab\_UQAM were all competitors in the TermEval 2020 shared task on the ACTER dataset \cite{rigouts2020termeval}. The TALN-LS2N approach is the winning solution on the English and French datasets, whereas NLPLab\_UQAM is the winning team on the Dutch corpus. While RACAI, NYU, and e-Terminology rely on feature engineering and statistical approaches, TALN-LS2N and NLPLab UQAM applied neural network-based models. After the competition, several new approaches were developed. For example, HAMLET is a novel supervised method inspired by traditional hybrid systems such as TermoStat. Furthermore, \cite{nugumanova2022nmf} combined probabilistic topic modeling (PTM) and non-negative matrix factorization (NMF), and the optimal combinations of NMF outperformed four baseline extraction methods (e.g., TF-IDF, RAKE, YAKE, and TextRank). All the described approaches were implemented on the ACTER dataset of version 1.2, which was first released by the shared task competition. However, none of them had considered the problem as a token-level sequence labeling task and had taken advantage of the Transformer-based models, until the novel approach from \cite{lang2021transforming,trantransformer} and \cite{trantransformer}.

\cite{lang2021transforming} proposed a comprehensive comparison between three Transformer-based ATE models operating on the sentence level: token classification (sequence labeling task), sequence classification, and sequence-to-sequence generation. This proved three points. First, the superiority of token classification in terms of performance, which, therefore, affirmed itself as the state-of-the-art (SOTA) with regard to the ACTER dataset. Second, the multilingual models perform better than the monolingual ones. Third, zero-shot learning shows good potential in extracting the candidate terms in new unseen languages and domains. \cite{tran2022ensembling} extended this research with an empirical evaluation of several monolingual and multilingual Transformer-based language models, including both masked (e.g., BERT and its variants) and autoregressive (e.g., XLNet) models, on the cross-domain ATE tasks with additional ensembling two best classifiers to optimize the Recall. Their approaches were applied to the ACTER dataset of version 1.5. However, there were not many changes to actual annotations, but major updates to how the annotations were presented.

Most deep learning based approaches also proved to be very competitive and surpassed the performance of most non-neural methods by a large margin on all languages and annotation types. \cite{lang2021transforming} and \cite{trantransformer} also showcased the potential of cross-lingual learning when considering ATE tasks as a token classification problem. These approaches currently represent the new state-of-the-art methods regarding most languages in the ACTER corpus.

\section{Challenges and Future directions}
\label{sec:discussion}

With the development of deep neural networks and the advent of recent Transformer-based models, deep learning-based ATE benefited significantly from the advances of pretrained embeddings in language models. We now have the opportunity to re-examine the term extraction task, the relative challenges, and the potential future directions, without having to involve complicated feature engineering.

Considering the scarcity of well-annotated and well-documented corpora, especially for less-resourced languages, the main objective in the field is to develop a robust approach for candidate terms detection that will be able to extract meaningful candidate terms from text sequences from a variety of domains and languages. This comes with the challenges addressed in the following subsections.
 
\subsection{Nested terms}

Nested terms refer to words or phrases that are both terms themselves and parts of longer terms. This means that a shorter term may appear within a larger term and vice versa. For example, in ACTER's English Corruption dataset, the word \textit{``confiscation''} is a standalone term that also appears as a part of other longer terms such as \textit{``confiscation of corruption proceeds''}, \textit{ ``confiscation of criminal assets''}, and \textit{``confiscation of the proceeds of crime''}. This makes the labeling harder, as the classifier needs to infer from the context whether a specific term is part of a longer term. This leads to the most common problem of current systems, which is that instead of predicting the whole multi-word term, they predict only a shorter gold standard term nested in the multi-word gold standard term. Vice versa, the model sometimes also generates incorrect predictions containing the correct nested terms. 

\subsection{Generalisability}

In the past decade, we have witnessed a growing data production involving different forms of data, including raw documents. However, building domain-specific corpora with manually annotated terms from the available raw texts is often error-prone, labor-intensive, time-consuming, and subjective\footnote{Subjectiveness emerges due to two major reasons: the low agreement among the annotators and the lack of consensus about an annotation protocol.}. As a consequence, annotated datasets, on which the models can be trained, are often scarce, limited, and usually do not cover many languages and domains. Therefore, it is necessary for the model to have the ability to adapt and extract terms from new languages and domains, which were not present in its training dataset. For this reason, one of the future challenges is to build models that will be able to transfer the knowledge between different domains and languages efficiently and to extract language- and domain-specific terminology even either without the language- or domain-specific training corpus, or with only a few available training examples.

With the recent advances in language models and the huge demand for real-world applications, we expect ATE to receive more attention from researchers in the future. Based on the comparative studies in this survey, we list some proposals for further directions that the research on term extraction could take.

\subsection{Ensembling} 

Deep neural networks based on pretrained language models have a strong predictive power on downstream tasks (e.g., term extraction), and can be fed raw data directly, without any feature engineering required. However, linguistic and statistical measures of termhoodness can still contribute to the research on ATE, by being added to neural models as additional features with the aim of boosting performance. The integration of models with different combinations of non-contextual features, contextual features from language models, and linguistics and statistical measures (e.g., termhood, unithood, C-value) should be investigated. Furthermore, \cite{tran2022ensembling} presented the potential of ensembling two best classifiers by simply performing the union operation on the candidate term lists of the best two classifiers. Further ensembling techniques should be employed to enhance the performance in Recall and by extension in F1-score.

\subsection{Scalability} 

Despite showing competitive performance in extracting good candidate terms, large pretrained neural language models are computationally expensive. Furthermore, new research is going in the direction of exponentially increasing the parameters of the model and the training data size in order to improve performance. While this has been shown to be a promising research direction, these models are usually too expensive for being used in production or in academia. Therefore, we believe that going in the opposite direction, i.e., making neural approaches more scalable, is also promising. We moreover propose to develop approaches with good trade-offs between performance and scalability. Model compression and pruning techniques are also good options to reduce the space and computation time required for model learning, without compromising the overall performance.

\subsection{Prompting} 
Prompt-based methods have been successfully applied to sentence-level few-shot learning tasks such as in NER tasks \cite{cui2021template}. Instead of replacing the language model head with the classifier head, prompt-based approaches induce the prediction from the language model by constructing a template. By re-using the masked language model objective, prompt-based approaches alleviate the problem of pre-training and fine-tuning discrepancy (caused by their different training objectives), as well as benefit the few-shot performance. Furthermore, to avoid the time-consuming issues of searching for appropriate templates, a template-free approach \cite{ma2021template}, which re-used the language model head and leads the model to predict label words through an Entity-oriented objective, was also introduced. However, few efforts have been made to apply these methods to the token-level learning tasks and ATE in specific. Therefore, we believe that there is still room for improvement in this direction.

\section{Conclusion}
\label{sec:conclusion}

Our survey summarizes the recent advances in the automatic term extraction task, covering both classic machine learning models based on feature engineering and novel neural network models that have yielded several important insights, especially with the advent of pretrained language models. We first surveyed different resources and presented a systematic list of well-annotated monolingual and multilingual corpora for term extraction from the past three decades. Then, we presented the first systematic summary of deep learning-based approaches and made a comparison of their performance in supervised term extraction to machine learning based approaches. Furthermore, we indicated all the metrics used in term extraction and categorized them based on the evaluation methodology and the scope of results. Incorporating the lessons learned from the existing techniques, we pointed out the current challenges and proposed some suggestions for further exploration in the term extraction research.

The main findings of this survey are that neural models generally outperform machine learning models based on feature engineering by a large margin and that Transformer-based cross-domain and cross-lingual models tend to perform really well, establishing new state-of-the-art methods on the benchmark ACTER dataset. When three types of Transformer-based ATE models were implemented to operate on the sentence level, the superiority of token classification in terms of performance affirmed itself even in cross-domain or cross-lingual in comparison with the monolingual setting. Despite monolingual language models (e.g, SloBERTa) on less-resourced languages (e.g., Slovenian) captured the information about terms than multilingual ones (e.g., multilingual BERT) if named entities are not included, multilingual ones outperform the others for extracting terms with entities and adding more data from other languages to the train set that matches the target language improves the predictive performance of the model. The zero-shot learning shows good potential in extracting the candidate terms in new unseen languages and domains.

\begin{acks}
The work was partially supported by the Slovenian Research Agency (ARRS) core research programme Knowledge Technologies (P2-0103), as well as the Ministry of Culture of the Republic of Slovenia through the project Development of Slovene in Digital Environment (RSDO). The first author was partly funded by Région Nouvelle-Aquitaine. This work has also been supported by the TERMITRAD (2020-2019-8510010) project funded by the Nouvelle-Aquitaine Region, France.
\end{acks}

\bibliographystyle{ACM-Reference-Format}
\bibliography{main}

\end{document}